\PassOptionsToPackage{unicode}{hyperref}
\PassOptionsToPackage{hyphens}{url}
\PassOptionsToPackage{dvipsnames,svgnames,x11names}{xcolor}
\documentclass[
  11pt,
]{article}
\usepackage{xcolor}
\usepackage[margin=1in]{geometry}
\usepackage{amsmath,amssymb}
\usepackage{iftex}
\ifPDFTeX
  \usepackage[T1]{fontenc}
  \usepackage[utf8]{inputenc}
  \usepackage{textcomp} 
\else 
  \usepackage{unicode-math} 
  \defaultfontfeatures{Scale=MatchLowercase}
  \defaultfontfeatures[\rmfamily]{Ligatures=TeX,Scale=1}
\fi
\usepackage{lmodern}
\ifPDFTeX\else
\fi
\IfFileExists{upquote.sty}{\usepackage{upquote}}{}
\IfFileExists{microtype.sty}{
  \usepackage[]{microtype}
  \UseMicrotypeSet[protrusion]{basicmath} 
}{}
\makeatletter
\@ifundefined{KOMAClassName}{
  \IfFileExists{parskip.sty}{%
    \usepackage{parskip}
  }{
    \setlength{\parindent}{0pt}
    \setlength{\parskip}{6pt plus 2pt minus 1pt}}
}{
  \KOMAoptions{parskip=half}}
\makeatother
\usepackage{color}
\usepackage{fancyvrb}

\DefineVerbatimEnvironment{Highlighting}{Verbatim}{commandchars=\\\{\}}
\newenvironment{Shaded}{}{}

\newcommand{\AttributeTok}[1]{\textcolor[rgb]{0.49,0.56,0.16}{#1}}

\newcommand{\BuiltInTok}[1]{\textcolor[rgb]{0.00,0.50,0.00}{#1}}

\newcommand{\ExtensionTok}[1]{#1}

\newcommand{\FunctionTok}[1]{\textcolor[rgb]{0.02,0.16,0.49}{#1}}

\newcommand{\NormalTok}[1]{#1}

\usepackage{longtable,booktabs,array}
\usepackage{calc} 
\usepackage{etoolbox}
\makeatletter
\patchcmd\longtable{\par}{\if@noskipsec\mbox{}\fi\par}{}{}
\makeatother
\IfFileExists{footnotehyper.sty}{\usepackage{footnotehyper}}{\usepackage{footnote}}
\makesavenoteenv{longtable}
\usepackage{graphicx}
\makeatletter
\newsavebox\pandoc@box
\newcommand*\pandocbounded[1]{
  \sbox\pandoc@box{#1}%
  \Gscale@div\@tempa{\textheight}{\dimexpr\ht\pandoc@box+\dp\pandoc@box\relax}%
  \Gscale@div\@tempb{\linewidth}{\wd\pandoc@box}%
  \ifdim\@tempb\p@<\@tempa\p@\let\@tempa\@tempb\fi
  \ifdim\@tempa\p@<\p@\scalebox{\@tempa}{\usebox\pandoc@box}%
  \else\usebox{\pandoc@box}%
  \fi%
}
\def\fps@figure{htbp}
\makeatother
\providecommand{\tightlist}{%
  \setlength{\itemsep}{0pt}\setlength{\parskip}{0pt}}
\usepackage{bookmark}
\IfFileExists{xurl.sty}{\usepackage{xurl}}{} 
\hypersetup{
  pdftitle={The Remarkable Effectiveness of Providing AI Agents with Natural Language Tools: A Replication Study},
  pdfauthor={A. Somma; I. Plante; F. Premji},
  colorlinks=true,
  linkcolor={blue},
  filecolor={Maroon},
  citecolor={Blue},
  urlcolor={blue},
  pdfcreator={LaTeX via pandoc}}

\title{The Remarkable Effectiveness of Providing AI Agents with Natural Language Tools: A Replication Study}
\usepackage{etoolbox}
\makeatletter
\providecommand{\subtitle}[1]{
  \apptocmd{\@title}{\par {\large #1 \par}}{}{}
}
\makeatother
\subtitle{Validating NLT Performance Across 14 Models}
\author{A. Somma, I. Plante, F. Premji \\[4pt] \normalsize Sage.is AI-UI}
\date{July 2, 2026}

\begin{document}
\maketitle
\begin{abstract}
This study independently replicates and extends the Natural Language Tools (NLT) framework of Johnson et al.~(2025), which questions the use of structured tool calling in large language model (LLM) agentic systems. We evaluated NLT across 14 models and 8,560 trials, adding newer frontier, reasoning, and open-weight models to the original set. The results confirm the core findings and add detail. NLT improves tool-calling accuracy by 14.9 percentage points overall (62.3\% versus 47.4\% structured) and reduces critical errors by 93\% (51 versus 755 errors). The gains depend on model capability: models without native tool calling, reasoning models, and smaller models gain substantially (+24.0pp to +43.1pp), while heavily optimized frontier models (GPT-5, Gemini 2.5 Pro) show smaller or reversed advantages. This matches recent analyses of reinforcement-learning-optimized tool use (Martinez, 2025). NLT also cuts token usage by 25.2\%. The reliability and efficiency advantages compound in recursive agentic workflows, where agents chain many tool calls across sub-agents: a structured failure triggers retries, fallback routing, and coordination overhead, while NLT avoids most of that cost at the source. This work makes three contributions: (1) the first independent validation of NLT using open-source tooling, (2) evidence that model capability moderates NLT's advantages (Chen et al., 2025; Zhang et al., 2025), and (3) a measurement of NLT's reliability benefit (93\% fewer errors), its most deployment-relevant property given the known fragility of structured tool calling. NLT is a practical alternative to structured tool calling, especially for production systems that value reliability over parseability.
\end{abstract}

\begin{center}\rule{0.5\linewidth}{0.5pt}\end{center}

\begin{figure}
\centering
\includegraphics[width=0.9\linewidth,height=\textheight,keepaspectratio,alt={xkcd \#2116: .NORM Normal File Format --- Randall Munroe, CC BY-NC 2.5 Why are we making the language model stop speaking language?}]{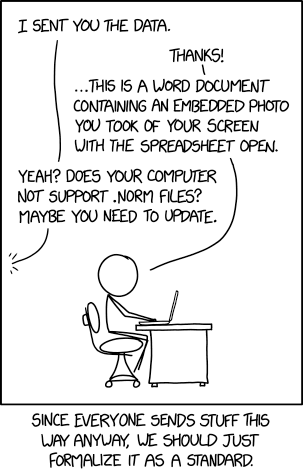}
\caption{\href{https://xkcd.com/2116/}{xkcd \#2116: .NORM Normal File Format} --- Randall Munroe, CC BY-NC 2.5 Why are we making the language model stop speaking language?}
\end{figure}

\begin{center}\rule{0.5\linewidth}{0.5pt}\end{center}

Note on accuracy reporting: Raw accuracy is calculated only over valid (non-error) trials. When a model produces errors on the majority of trials (e.g., 76 out of 80), the few remaining responses may yield misleadingly high accuracy figures, leading to survivorship bias. In this study, any condition with errors on 70 or more of 80 trials is treated as having 0\% accuracy, reflecting operational failure rather than selective success.


\begin{center}\rule{0.5\linewidth}{0.5pt}\end{center}

\section{1. Introduction}\label{introduction}

\subsection{1.1 Background}\label{background}

Generative AI, beginning around 2020 with models like GPT-3, changed natural language processing. Large language models (LLMs) can now produce coherent text and handle complex inference tasks. Early uses centered on text generation, but the work soon extended to agent-based systems, where LLMs use external tools to achieve goals such as retrieving data or taking actions. Tool calling lets an LLM invoke functions or APIs, and it underpins most agentic systems today. It is usually implemented with structured formats such as JSON schemas, chosen for reliability and ease of parsing. Agentic systems grew quickly through late 2025 and into 2026, across enterprise workflow orchestration and consumer-facing autonomous agents such as OpenClaw (Steinberger, 2026). At that scale, reliable tool calling is a production requirement, not just a research question.

We propose that structured tool calling introduces a cognitive trade-off that impairs performance on domain-specific problem-solving tasks. This view is consistent with the literature on format constraints in LLMs. The limitation is not that LLMs cannot follow JSON schemas; modern models are strong at code generation (Chen et al., 2021). Rather, following a schema appears to redirect the model's representational resources away from the primary task, so format requirements compete with task instructions for cognitive bandwidth. Work on prompt sensitivity in LLMs points the same way (Reynolds \& McDonell, 2021).

We hypothesize that schema formatting compels the model to draw from distinct segments of its learned distribution. JSON generation patterns are primarily acquired from coding corpora, whereas tasks such as customer service or mental wellness originate from different domains. This distributional mismatch fragments the model's attention and diminishes its effectiveness in core reasoning tasks, even when the output remains syntactically valid.

The cognitive load of maintaining format compliance may engage what Kahneman (2011) describes as System 2 thinking (deliberate, effortful processing) at the expense of System 1's intuitive task understanding. In LLMs, this manifests as competition between instruction following and format adherence, particularly pronounced in models not extensively fine-tuned for structured outputs (Weston et al., 2023). This distributional mismatch may be compounded by the integration of expert systems, in which domain-specific knowledge must be mapped to generic format constraints.

Models also turn over quickly, which raises the need for flexible tool-calling methods that reduce deployment risk. Recent work keeps exploring natural language approaches: ToolFlow (Chen et al., 2025) uses dialogue synthesis to improve tool-calling performance, and CallNavi (Zhang et al., 2025) is an empirical study of function-calling challenges. Both show that natural language tool interfaces remain relevant and that they need careful evaluation. The spread of autonomous agent frameworks, including the widely used OpenClaw project, which routes tool calls through LLMs such as Claude, GPT, GLM, and DeepSeek, makes it practically important to know when natural language interfaces beat structured ones.

Johnson et al.~(2025) showed that replacing programmatic JSON tool calling with natural language (NLT) significantly improved LLM tool-calling accuracy. They reported an 18.4 percentage point gain across 10 models and 6,400 trials, along with lower variance and token savings.

These findings question the prevailing paradigm of structured tool calling by demonstrating that format constraints constitute a significant, often overlooked bottleneck in agentic system performance. Adoption of NLT has the potential to alter how developers implement tool calling in production environments. The rapid introduction of new models underscores the need for adaptable tool-calling methods to mitigate deployment risks. This replication of Johnson et al.~(2025) aims to validate these claims using a broader set of models, including frontier, reasoning, and open-weight models introduced since the original study. The results presented will inform a planned follow-up study designed to expand the experimental scope beyond the original conditions.

\subsection{1.2 Motivation for Replication}\label{motivation-for-replication}

Replication studies matter for scientific progress in machine learning, especially given ongoing concerns about reproducibility in AI research (Pineau et al., 2020). We undertook this independent replication for several reasons:

1. Validate Core Findings: Johnson et al.~(2025) reported substantial gains with NLT (+18.4pp), challenging the dominant structured tool-calling paradigm. Independent validation using a separate codebase and evaluation framework tests whether these results generalize beyond the original implementation.

2. Assess Generalizability: The original study evaluated 13 models available in October 2025. Since then, new frontier models (GPT-5, Claude Sonnet 4, Gemini 2.5 Pro) and reasoning models (DeepSeek-R1) have emerged. Testing these newer models reveals whether NLT's advantages persist as model capabilities evolve.

3. Test Robustness: Implementation details, prompt variations, and API differences can significantly affect results (Zhao et al., 2021). By implementing NLT from scratch with different tooling, we test the robustness of the original findings to implementation variations.

4. Provide Open-Source Tooling: The field benefits from accessible evaluation frameworks. Our open-source implementation enables continued NLT research and lowers barriers to entry for other researchers.

5. Identify Boundary Conditions: By testing a broader range of models than the original study, we can identify where NLT's advantages may diminish or reverse. Thus is crucial information for practitioners deciding when to adopt NLT versus structured approaches.

Gundersen et al.~(2023) emphasize that independent replication is essential for distinguishing robust findings from implementation artifacts or publication bias. Our study contributes to this scientific process by providing transparent methodology, open-source code, and comprehensive results that others can verify and build upon.

\subsection{1.3 Scope and Limitations}\label{scope-and-limitations}

Our replication focuses on the core experimental conditions from Johnson et al.~(2025).

\begin{itemize}
\tightlist
\item
  Single-turn, parameterless tool selection
\item
  Two scenarios: customer service (``Alex'') and mental health (``Sage'').
\item
  Exact-match evaluation with 5 replicates per input
\item
  Comparison of NLT vs structured tool calling approaches
\end{itemize}

Limitations we acknowledge:

\begin{itemize}
\tightlist
\item
  Model Availability: Some models from the original study are unavailable, producing potential differences in performance across tested models. Two of our 14 models have partial data (Gemini 2.5 Pro and Qwen3-VL) due to API availability during the evaluation window.
\item
  API Differences: API implementations may differ from the original study. For example, during our tests, API latency varied up to 120 ms across providers, affecting the response time and potentially the accuracy of results.
\item
  Temporal Effects: Model capabilities may have changed since the original evaluation, revealing ongoing optimization and updates.
\item
  Infrastructure: Different inference infrastructure may affect results, given variability in processing speeds and network conditions.
\item
  We do not replicate multi-turn interactions. We also do not replicate parameterized tool calls.
\end{itemize}

\subsection{1.4 Contributions}\label{contributions}

\begin{enumerate}
\def\labelenumi{\arabic{enumi}.}
\tightlist
\item
  Independent validation of NLT's core claims using open-source tooling
\item
  Extended evaluation to 14 models, including 5 not in the original study
\item
  Detailed per-model analysis revealing a capability-dependent pattern in NLT gains
\item
  Open-source framework for continued NLT research and evaluation
\item
  Reproducibility artifacts, including all prompts, inputs, and raw results
\item
  Identification of boundary conditions for a planned follow-up study
\end{enumerate}

\begin{center}\rule{0.5\linewidth}{0.5pt}\end{center}

\section{2. Methodology}\label{methodology}

\subsection{2.1 Implementation}\label{implementation}

An independent Python-based evaluation harness was developed that implements the NLT framework as described by Johnson et al.~(2025). The implementation includes:

Core Components:

\begin{itemize}
\tightlist
\item
  \href{https://github.com/Sage-is/NLT-Replication-Study/blob/develop/src/nlt/core/evaluator.py}{evaluator.py}: Trial execution and metrics computation
\item
  \href{https://github.com/Sage-is/NLT-Replication-Study/blob/develop/src/nlt/core/parser.py}{parser.py}: YES/NO parsing for NLT and tool\_calls extraction for structured
\item
  \href{https://github.com/Sage-is/NLT-Replication-Study/blob/develop/src/nlt/core/client.py}{client.py}: API client using urllib (no external dependencies)
\item
  \href{https://github.com/Sage-is/NLT-Replication-Study/blob/develop/src/nlt/core/scenarios.py}{scenarios.py}: Alex and Sage scenarios with tool schemas
\end{itemize}

Key Implementation Decisions:

\begin{itemize}
\tightlist
\item
  Pure Python with no external API libraries (urllib only)
\item
  OpenAI-compatible function calling schema for a structured approach
\item
  Regex-based parsing for NLT YES/NO extraction
\item
  Exact-match grading with no partial credit
\item
  5 independent replicates per input (parallel API calls)
\end{itemize}

\subsection{2.2 Experimental Design}\label{experimental-design}

We replicated the original 2 \(\times\) 2 \(\times\) 2 factorial design. Each factor in this design serves a specific purpose: the approach tests the contrast between NLT and structured tool calling, the scenario differentiates between customer service (Alex) and mental health (Sage), and the perturbation examines prompt brittleness by using semantically equivalent but stylistically different instructions. This framework allows us to thoroughly evaluate the robustness and accuracy of the models across different conditions.

\begin{itemize}
\tightlist
\item
  Approach: NLT vs Structured Tool Calling
\item
  Scenario: Alex (customer service) vs Sage (mental health)
\item
  Perturbation: Non-perturbed vs Perturbed prompts
\end{itemize}

Per-model trial count (for models with complete data):

\begin{itemize}
\tightlist
\item
  2 approaches \(\times\) 2 scenarios \(\times\) 16 inputs \(\times\) 2 perturbations \(\times\) 5 replicates = 640 trials per model
\item
  Total: 8,560 trials across 14 models (107 aggregated entries)
\end{itemize}

Comparison to Original:

\begin{itemize}
\tightlist
\item
  Key differences: Model selection (newer models), API implementations, and evaluation timeframe
\end{itemize}

Per-model trial count (for models with complete data):

\begin{itemize}
\tightlist
\item
  2 approaches \(\times\) 2 scenarios \(\times\) 16 inputs \(\times\) 2 perturbations \(\times\) 5 replicates = 640 trials per model
\item
  Original study: 6,400 trials across 10 models
\end{itemize}

\subsection{2.3 Scenarios and Tool Definitions}\label{scenarios-and-tool-definitions}

\subsubsection{2.3.1 NLT}\label{nlt}

We used identical scenarios, tool descriptions and simulated user inputs from Johnson et al.~(2025):

Example Alex (Customer Service) Natural Language Tools and Scenario Description:

You are an assistant to Alex, an AI customer service agent who handles bookings for a music venue called ``Yes! Music''. You will be given a message between Alex and a customer. They are texting one another.\\
Your mission is to identify if any of the following topics have been brought up\ldots{} (Further detail can be seen in \href{https://github.com/Sage-is/NLT-Replication-Study/blob/2aa826401e18eaf5aae8bb17db34a1a09eb444f5/src/nlt/data/scenarios.py\#L6}{scenarios.py line 6})

Additionally, the Alex agent was allowed to select what tools to access by responding yes or no to the following list of tools, or explaining Thinking or stating the Assessment finished:

Thinking: (insert\_thinking)\\
Recap of previous conversation -- YES/NO\\
Website information -- YES/NO\\
Recent social media posts -- YES/NO\\
Available discounts -- YES/NO\\
List of upcoming events -- YES/NO\\
Past Purchases -- YES/NO\\
Talk to a Human -- YES/NO\\
Assessment finished.

Example Sage (Mental Health) Natural Language Tools and Scenario Description:

You are an assistant to Sage, an AI mental health specialist. You will be given a message between Sage and their client. They are texting one another.\\
Your mission is to identify if any of the following topics have been brought up\ldots{} (further details can be seen in \href{https://github.com/Sage-is/NLT-Replication-Study/blob/2aa826401e18eaf5aae8bb17db34a1a09eb444f5/src/nlt/data/scenarios.py\#L92}{scenarios.py line 92})

Additionally, the Sage agent was allowed, similar to Alex, to select what tools to access by responding yes or no to the following list of tools, or explaining Thinking or stating the Assessment finished:

Thinking: (insert\_thinking)\\
Most Recent Conversation -- YES/NO\\
Psychometric Quizzes -- YES/NO\\
Sage Website Information -- YES/NO\\
Sage Technology -- YES/NO\\
Sage Company Info -- YES/NO\\
Sage Social Media -- YES/NO\\
End Conversation -- YES/NO\\
Safety Call -- YES/NO\\
Assessment finished.

\subsubsection{2.3.2 Structured}\label{structured}

We used industry-standard structured tooling based on the Johnson et al.~(2025) scenarios and toolings listed in their study. An example of Alex's structured tool calling scenario can be seen on \href{https://github.com/Sage-is/NLT-Replication-Study/blob/2aa826401e18eaf5aae8bb17db34a1a09eb444f5/src/nlt/data/scenarios.py\#L73}{line 73 of scenarios.py}.

\subsection{2.4 Model Selection}\label{model-selection}

Original Study's Approach:\\
Johnson et al.~(2025) evaluated 13 models spanning open and closed families, selected based on popularity via the OpenRouter leaderboard. They divided models into:

\begin{itemize}
\tightlist
\item
  Core set (10 models): Tested on both NLT and Structured approaches
\item
  Auxiliary set (3 models): DeepSeek R1-0528, GPT-OSS-120B, GPT-OSS-20B --- tested only on NLT due to ``limited tool calling capabilities at evaluation time.''
\end{itemize}

Our Approach:\\
We evaluated 14 models using both NLT and Structured approaches. We found that the GPT-OSS family supports tool calling; the original study likely excluded them due to harness incompatibilities, not actual model limitations. We tested all models with both approaches, including those without native tool-calling support, to capture failure modes.

Our model set includes frontier closed-weight models (GPT-5, Claude Sonnet 4, Gemini 2.5 Pro), mid-tier models (Gemini 2.0 Flash, Gemini 2.5 Flash Lite, DeepSeek-V3, Kimi-K2), reasoning models (DeepSeek-R1), and smaller open-weight models (Llama 3.1 8B, Mistral 7B).

Evaluated Models (14):

\begin{enumerate}
\def\labelenumi{\arabic{enumi}.}
\tightlist
\item
  Anthropic/claude-sonnet-4
\item
  Deepseek/deepseek-chat-v3-0324
\item
  Deepseek/deepseek-r1
\item
  Google/gemini-2.0-flash-001
\item
  Google/gemini-2.5-flash-lite
\item
  Google/gemini-2.5-pro
\item
  Meta-llama/llama-3.1-8b-instant
\item
  Mistralai/mistral-7b-instruct (No native tool calling)
\item
  Moonshotai/kimi-k2
\item
  Openai/gpt-5
\item
  Openai/gpt-5-nano
\item
  Openai/gpt-oss-120b:free (Originally ``auxiliary'')
\item
  Openai/gpt-oss-20b:free (Originally ``auxiliary'')
\item
  Qwen/qwen3-vl-235b-a22b-thinking
\end{enumerate}

Data Completeness:

12 of 14 models have complete data across all 8 conditions. Two models have partial data: Google Gemini 2.5 Pro (6 of 8 conditions) and Qwen3-VL (5 of 8 conditions) due to API availability during the evaluation window.

\subsection{2.5 Prompt Design}\label{prompt-design}

We replicated the original scenario prompts with minimal adaptations for API compatibility:

\begin{itemize}
\tightlist
\item
  NLT Scenario Prompts: Natural language tool list with YES/NO format instructions
\item
  Structured Scenario Prompts: Function schemas passed via API with system prompt
\end{itemize}

\subsection{2.6 Evaluation Metrics}\label{evaluation-metrics}

\begin{itemize}
\tightlist
\item
  Accuracy: Proportion of exact matches (predicted tools = expected tools)
\item
  Corrected Accuracy: When a condition produces errors on \$\geq\$70 of 80 trials, we treat it as having an effective accuracy of 0\%. Raw accuracy over only surviving trials introduces a survivorship bias --- a model that errors on 76/80 trials but gets the remaining 4 correct would report 100\% accuracy, misrepresenting what is effectively a catastrophic failure. This correction affects 8 of 107 entries (2 models: all Qwen-structured conditions and all Mistral-structured conditions, 4 each).
\item
  Variance: Sample variance across replicates
\item
  Token Usage: Input, output, and total tokens per trial
\item
  Error Rate: Proportion of API errors or parsing failures
\end{itemize}

Our original evaluator can be reviewed in our project's \href{https://github.com/Sage-is/NLT-Replication-Study/blob/2aa826401e18eaf5aae8bb17db34a1a09eb444f5/src/nlt/core/evaluator.py\#L1}{evaluator.py} source code.

\subsection{2.7 Data Collection}\label{data-collection}

API Access:

\begin{itemize}
\tightlist
\item
  Open-weight models via various inference providers hosted using Startr.LLC's \href{http://sage.is/}{Sage.is} AI-UI
\item
  Closed-weight models via native provider APIs accessed through Startr.LLC's \href{http://sage.is/}{Sage.is} AI-UI
\item
  All models accessed with default parameters (temperature=1.0, top\_p=1.0)
\end{itemize}

Quality Control:

\begin{itemize}
\tightlist
\item
  API errors are retried until a clean response is obtained
\item
  All raw outputs logged for manual inspection
\item
  Parser validated against ground truth labels.
\end{itemize}

\begin{center}\rule{0.5\linewidth}{0.5pt}\end{center}

\section{3. Results}\label{results}

\subsection{3.1 Overall Accuracy}\label{overall-accuracy}

\begin{figure}
\centering
\includegraphics[width=0.9\linewidth,height=\textheight,keepaspectratio,alt={Charts generated by generate-charts.py}]{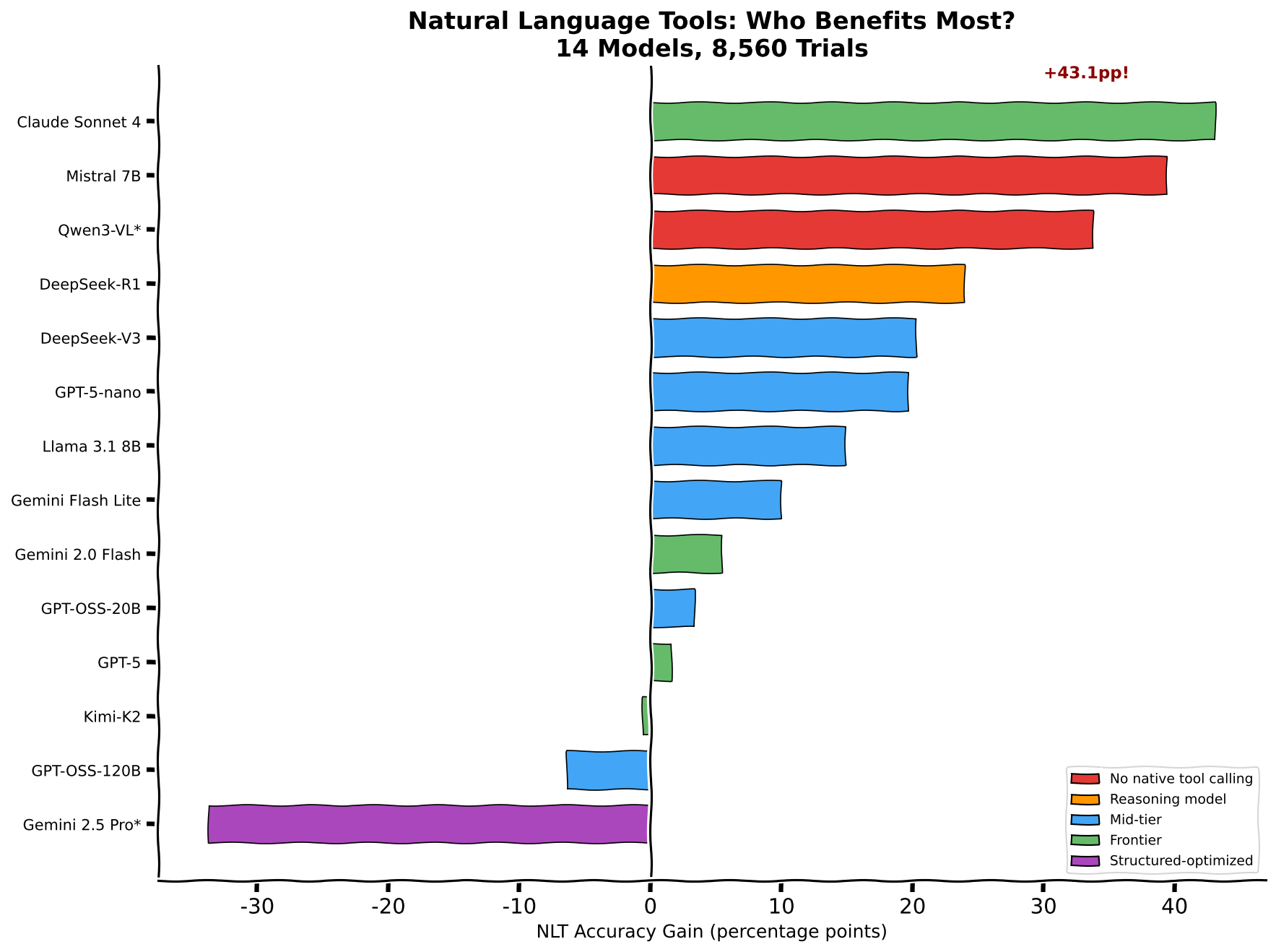}
\caption{Charts generated by \href{https://github.com/Sage-is/NLT-Replication-Study/blob/develop/assets/generate-charts.py}{generate-charts.py}}
\end{figure}

Replication Results (14 models, 107 entries, 8,560 trials):

\begin{itemize}
\tightlist
\item
  NLT accuracy: 62.3\% vs Structured accuracy: 47.4\% (corrected).
\item
  \(\Delta\) = +14.9pp overall gain.
\item
  Total Errors: NLT (51) vs Structured (755). Structured approach failure rates were dramatically higher --- a 93\% error reduction with NLT.
\item
  NLT outperformed structured approaches in 11 of 14 models.
\end{itemize}

\includegraphics[width=0.9\linewidth,height=\textheight,keepaspectratio]{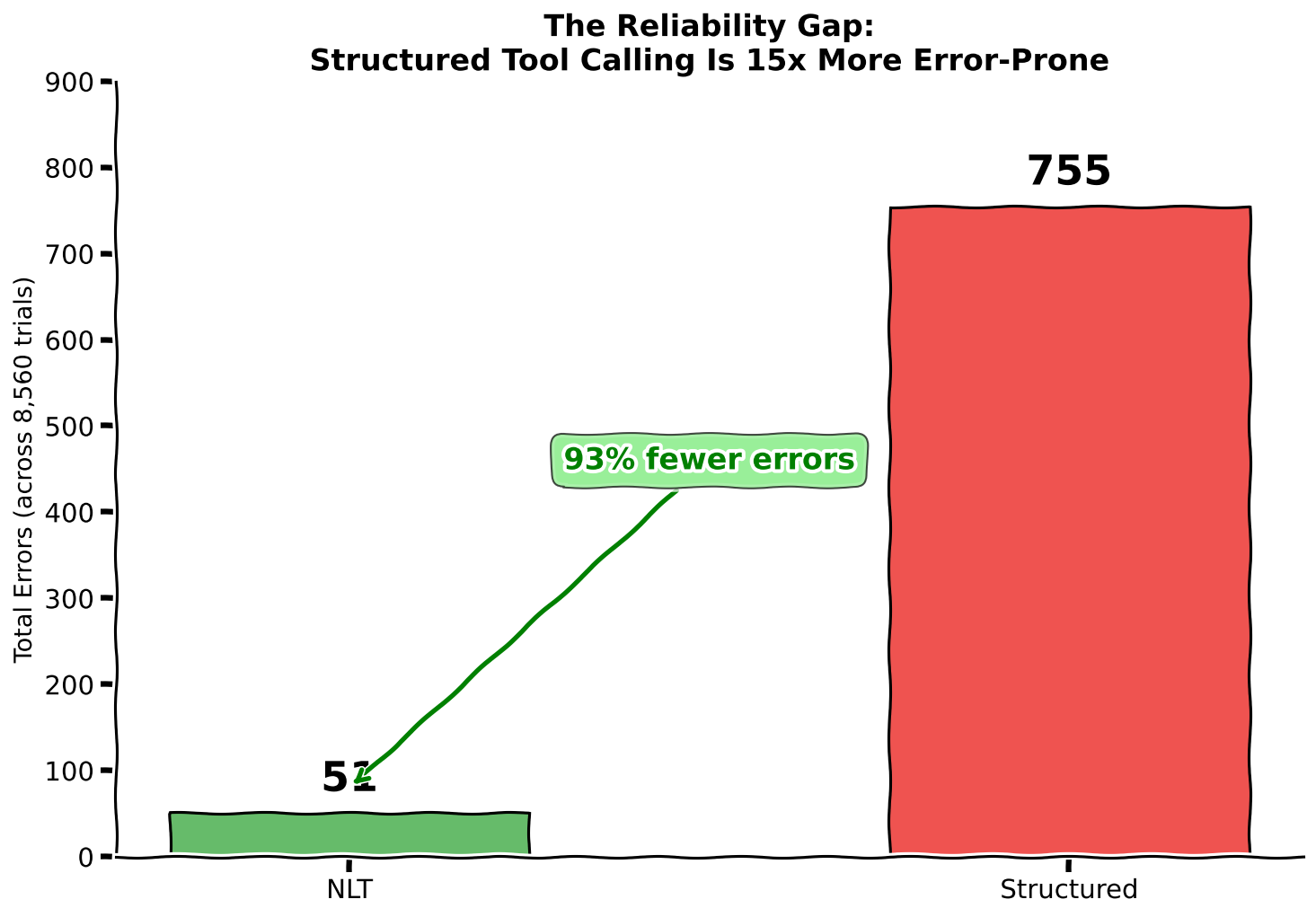}

Note: Structured accuracy uses corrected figures. Raw structured accuracy was 51.9\%, but this is inflated by survivorship bias in conditions where nearly all trials errored (see Section 2.6). For example, Qwen's structured Alex non-perturbed condition errored on 76 of 80 trials but reported 100\% accuracy on the 4 surviving responses. We correct such conditions (errors \(\geq\) 70/80) to 0\% accuracy.

Comparison to Original Study:

\begin{itemize}
\tightlist
\item
  Original overall gain: +18.4pp across 10 models and 6,400 trials.
\item
  Our replication: +14.9pp across 14 models and 8,560 trials (corrected).
\item
  The effect is confirmed. The reduced gain can be explained by the inclusion of frontier models (GPT-5, Gemini 2.5 Pro) that have been heavily optimized for structured tool calling, showing near-parity or reversed gains. Differences are also likely due to model selection, lack of access to the original testing harness, and improvements in structured tool-calling support in newer model generations.
\end{itemize}

\subsection{3.2 Per-Model Performance}\label{per-model-performance}

Model Performance (sorted by NLT gain):

\begin{itemize}
\tightlist
\item
  Anthropic/claude-sonnet-4: +43.1pp Gain (NLT 61.9\% / Structured 18.8\%)

  \begin{itemize}
  \tightlist
  \item
    Note: Largest NLT gain in our study. Claude's structured accuracy was exceptionally low despite being a frontier model.
  \end{itemize}
\item
  Mistralai/mistral-7b-instruct: +39.4pp Gain (NLT 39.4\% / Structured 0.0\%)

  \begin{itemize}
  \tightlist
  \item
    Note: Mistral failed completely on structured (320 errors). No native tool-calling support.
  \end{itemize}
\item
  Deepseek/deepseek-r1: +24.0pp Gain (NLT 55.0\% / Structured 31.0\%)

  \begin{itemize}
  \tightlist
  \item
    Note: The reasoning model shows large NLT gains, suggesting that chain-of-thought interferes with structured output.
  \end{itemize}
\item
  Deepseek/deepseek-chat-v3-0324: +20.3pp Gain (NLT 90.0\% / Structured 69.7\%)
\item
  Openai/gpt-5-nano: +19.7pp Gain (NLT 79.1\% / Structured 59.4\%)
\item
  Meta-llama/llama-3.1-8b-instant: +14.9pp Gain (NLT 47.8\% / Structured 32.9\%)
\item
  Google/gemini-2.5-flash-lite: +10.0pp Gain (NLT 73.1\% / Structured 63.1\%)
\item
  Google/gemini-2.0-flash-001: +5.5pp Gain (NLT 85.0\% / Structured 79.5\%)
\item
  Openai/gpt-oss-20b:free: +3.4pp Gain (NLT 42.7\% / Structured 39.3\%)
\item
  Openai/gpt-5: +1.6pp Gain (NLT 81.9\% / Structured 80.3\%)

  \begin{itemize}
  \tightlist
  \item
    Note: Near-parity. GPT-5's structured tool calling is highly optimized.
  \end{itemize}
\item
  Moonshotai/kimi-k2: -0.6pp Loss (NLT 67.2\% / Structured 67.8\%)
\item
  Openai/gpt-oss-120b:free: -6.4pp Loss (NLT 42.6\% / Structured 49.0\%)
\item
  Qwen/qwen3-vl-235b-a22b-thinking: +33.8pp Gain (NLT 33.8\% / Structured 0.0\% corrected)

  \begin{itemize}
  \tightlist
  \item
    Note: Partial data (1 NLT entry vs 4 structured). Structured had 307 errors out of 320 trials. Raw accuracy was 60.7\% due to survivorship bias --- the few non-error responses happened to be correct. We correct to 0\% as all 4 conditions had \(\geq\) 73 errors out of 80 trials, representing operational failure.
  \end{itemize}
\item
  Google/gemini-2.5-pro: -33.7pp Loss (NLT 48.3\% / Structured 82.1\%)

  \begin{itemize}
  \tightlist
  \item
    Note: Partial data (3 entries each). Gemini 2.5 Pro is the strongest outlier, favouring structured tool calling.
  \end{itemize}
\end{itemize}

\includegraphics[width=0.9\linewidth,height=\textheight,keepaspectratio]{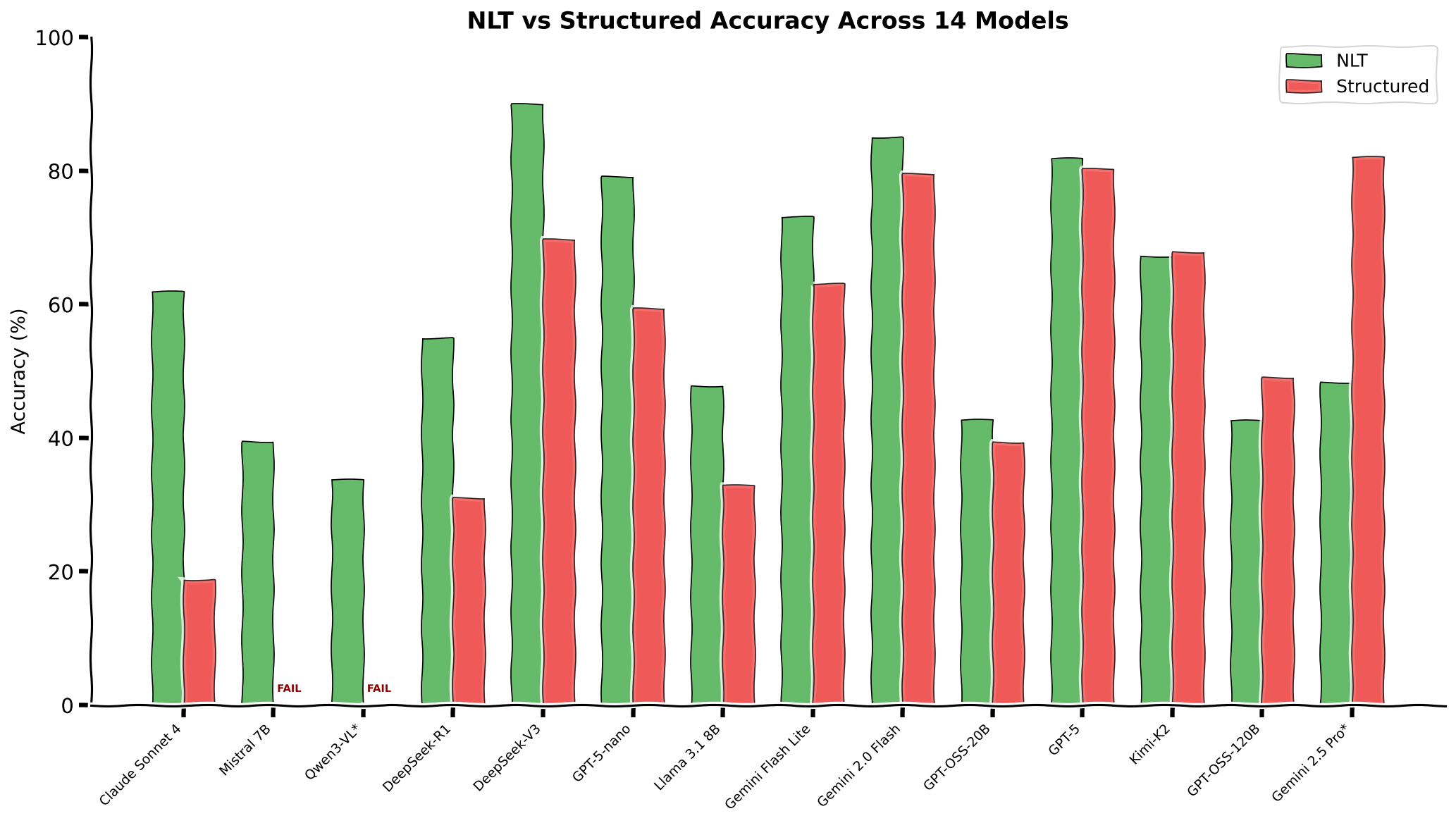}

\subsection{3.3 Variance Results}\label{variance-results}

Unlike the original study, which reported a 70\% reduction in variance with NLT, our results show comparable variance between the approaches. This is primarily because failures in the structured approach (755 errors, 0\% accuracy) compress the measured variance. When a model fails entirely on structured (e.g., Mistral with 0\% accuracy and zero variance), it artificially deflates the aggregate structured variance. The variance comparison is therefore less meaningful than the error rate comparison.

Comparison to Original:

\begin{itemize}
\tightlist
\item
  Original: Claimed significant variance reduction (70\%) with NLT.
\item
  Our replication: Mixed results. NLT variance (0.1913) is slightly higher than structured (0.1702), but this is confounded by systematic structured failures.
\item
  The primary differentiator in our study was reliability (error rate) rather than the variance of successful outputs.
\end{itemize}

\subsection{3.4 Perturbation Robustness}\label{perturbation-robustness}

To evaluate the fragility of each approach, we tested models with perturbed system prompts. Unlike input noise, such as typos in user messages, these perturbations used semantically equivalent but stylistically different instructions. The perturbed prompts used more verbose and complex language to describe the same tasks and tools, for example, changing ``Your mission is to identify\ldots{}'' to ``Serving as Alex's dedicated support assistant, you collaborate with\ldots{}''. This tests the model's sensitivity to prompt phrasing, which is a known issue in structured tool calling.

Non-perturbed Results (n=26 NLT, n=27 Structured):

\begin{itemize}
\tightlist
\item
  NLT: 62.4\% vs Structured: 46.1\% (corrected)
\item
  Gain: +16.4pp
\end{itemize}

Perturbed Results (n=26 NLT, n=28 Structured):

\begin{itemize}
\tightlist
\item
  NLT: 62.2\% vs Structured: 48.8\% (corrected)
\item
  Gain: +13.4pp
\end{itemize}

Comparison to Original:

\begin{itemize}
\tightlist
\item
  Original non-perturbed gain: +21.2pp
\item
  Original perturbed gain: +15.4pp
\item
  Our replication: NLT consistently outperformed structured approaches in both conditions, with corrected gains of +16.4pp (non-perturbed) and +13.4pp (perturbed). NLT accuracy was stable across perturbation states (62.4\% vs 62.2\%), demonstrating robustness to prompt phrasing. Structured accuracy also remained relatively stable after correction (46.1\% vs 48.8\%), suggesting the perturbation effect is less pronounced in our model set than in the original study.
\end{itemize}

\subsection{3.5 Domain Comparison (Alex vs Sage)}\label{domain-comparison-alex-vs-sage}

NLT gain was larger for Sage (+16.1pp) than Alex (+13.4pp), suggesting NLT provides greater benefit in the more complex mental health domain where structured approaches struggle most. Error counts were nearly equal across domains (NLT: 11 vs 40; Structured: 374 vs 381), indicating structured fragility was domain-independent.

Alex (Customer Service):

\begin{itemize}
\tightlist
\item
  NLT Accuracy: 66.2\% (n=26, 11 errors)
\item
  Structured Accuracy: 52.7\% (n=27, 374 errors, corrected)
\item
  Gain: +13.4pp
\end{itemize}

Sage (Mental Health):

\begin{itemize}
\tightlist
\item
  NLT Accuracy: 58.5\% (n=26, 40 errors)
\item
  Structured Accuracy: 42.4\% (n=28, 381 errors, corrected)
\item
  Gain: +16.1pp
\end{itemize}

Comparison to Original:

\begin{itemize}
\tightlist
\item
  Original NLT Accuracy: 87.5\%
\item
  Original Structured Accuracy: 69.1\%
\item
  Gain: 18.4pp
\end{itemize}

Across both domains (Alex and Sage), our replication confirms the original result: both benefit from NLT over structured tool calling. Overall accuracy, in either approach, is higher in the customer service domain (Alex) than in the mental health domain (Sage).

\subsection{3.6 Token Usage}\label{token-usage}

Across the board, NLT was significantly more token-efficient, reducing token usage by 25.2\%.

Token Reduction:

\begin{itemize}
\tightlist
\item
  NLT: 3,384,196 tokens
\item
  Structured: 4,522,651 tokens
\item
  Reduction: 25.2\%
\end{itemize}

\includegraphics[width=0.9\linewidth,height=\textheight,keepaspectratio]{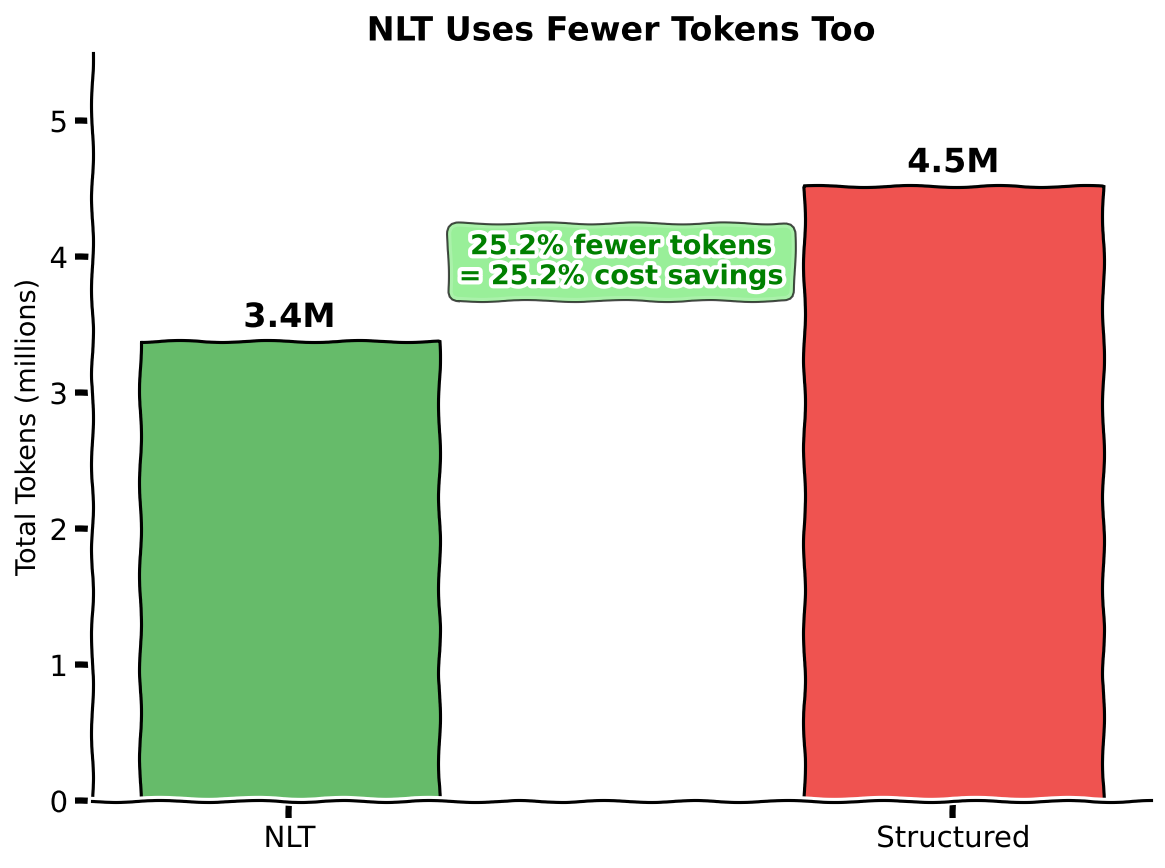}

Comparison to Original:

\begin{itemize}
\tightlist
\item
  NLT average: 905 tokens
\item
  Structured average: 1319 tokens
\item
  Reduction: 31.4\%
\end{itemize}

We confirmed that NLT is significantly more token-efficient, validating the original finding of reduced overhead. The slightly lower reduction may reflect model-specific differences in response verbosity across our expanded model set.

\subsection{3.7 Additional Observations}\label{additional-observations}

Catastrophic Failures in Structured Mode:
\textgreater{} Two models (Mistral-7B-Instruct, Qwen3-VL) collapsed completely with structured outputs: 0\% accuracy and hundreds of validation errors, mostly from failing to generate valid JSON. With NLT, both worked (39.4\% and 33.8\% accuracy) and produced zero validation errors. Claude Sonnet showed the same pattern unexpectedly, reaching only 18.8\% structured accuracy despite being a frontier model. This suggests that even capable models can have weak structured tool-calling implementations depending on the API integration path, and it matches research showing that format constraints create performance bottlenecks, particularly for models not heavily fine-tuned for structured outputs (Reynolds \& McDonell, 2021).

Frontier Model Convergence:
\textgreater{} GPT-5 and Gemini 2.0 Flash showed near-parity between NLT and structured approaches (+1.6pp and +5.5pp, respectively), with both achieving over 80\% accuracy in both modes. This suggests that highly optimized frontier models may have narrowed the distribution mismatch that NLT exploits through extensive tool-calling fine-tuning. Gemini 2.5 Pro went further, showing a strong structured advantage (-33.7pp), indicating that some models have been specifically optimized for structured output to the point where NLT is disadvantageous. This convergence demonstrates wider trends in model development, in which frontier models increasingly exhibit ``sparks of artificial general intelligence,'' including sophisticated tool-use capabilities (Bubeck et al., 2023). Recent analysis by Martinez (2025) suggests that reinforcement learning optimization for tool use has become a standard technique in frontier models, potentially explaining this proficiency in structured output.

Reasoning Model Penalty:
\textgreater DeepSeek-R1, a reasoning model, showed a large NLT gain (+24.0pp). Reasoning models generate extended chain-of-thought sequences before producing output. When constrained to structured formats, this reasoning process may conflict with those formats, degrading accuracy. NLT's free-form output naturally accommodates the reasoning trace. This finding extends research on chain-of-thought prompting, which shows that reasoning benefits from flexible output formats (Wei et al., 2022; Zhou et al., 2023) and aligns with recent work by Li et al.~(2025) on improving function calling and reasoning in LLMs, which similarly identifies conflicts between structured outputs and extended reasoning traces. The conflict between structured output requirements and extended reasoning may explain why reasoning models benefit disproportionately from NLT.

Model Size and NLT Benefit:
\textgreater Smaller models (Llama 3.1 8B, Mistral 7B) showed larger relative gains from NLT (+14.9pp and +39.4pp respectively), while larger frontier models showed smaller or reversed gains. This pattern aligns with scaling law research suggesting that larger models develop more sophisticated capabilities, including better compliance with formats (Kaplan et al., 2020). However, even frontier models maintained NLT's error-rate advantage, suggesting that reliability improvements may be orthogonal to accuracy convergence.

Domain Complexity Effects:
\textgreater The mental health scenario (Sage) showed larger NLT gains (+16.1pp) than the customer service scenario (+13.4pp), suggesting that NLT provides greater benefit in more complex domains where structured approaches struggle most. This corresponds to findings that complex tasks benefit more from flexible output formats that don't constrain reasoning processes (Zhou et al., 2023).

Our results can be reviewed in our project's \href{https://github.com/Sage-is/NLT-Replication-Study/tree/develop/results}{results} folder, as well as a table of of our \href{https://github.com/Sage-is/NLT-Replication-Study/blob/develop/aggregated_results.csv}{aggregated results}.

\begin{center}\rule{0.5\linewidth}{0.5pt}\end{center}

\section{4. Analysis and Discussion}\label{analysis-and-discussion}

\subsection{4.1 Validation of Core Findings}\label{validation-of-core-findings}

Confirmed:

\begin{itemize}
\tightlist
\item
  NLT outperforms structured approaches in accuracy among the majority of models (11 of 14 models show gains after correcting for survivorship bias).
\item
  NLT is significantly more robust to API failures. Models like Mistral and Qwen failed completely (0\% effective accuracy) with structured tool calling but performed reasonably well with NLT (39.4\% and 33.8\%, respectively). Claude Sonnet 4, despite being a frontier model, achieved only 18.8\% structured accuracy, compared with 61.9\% with NLT.
\item
  NLT reduces token usage (25.2\% reduction), consistent with the original finding (31.4\%).
\item
  Alex (customer service) yields higher accuracy than Sage (mental health) across both approaches.
\end{itemize}

Partially Confirmed:

\begin{itemize}
\tightlist
\item
  Variance reduction was not observed in our study. NLT variance (0.1913) was slightly higher than the structured variance (0.1702), though this comparison is confounded by systematic structured failures that compress the measured variance.
\end{itemize}

Not Confirmed:

\begin{itemize}
\tightlist
\item
  Universal NLT advantage. Three models showed structured advantages after correction (Gemini 2.5 Pro, GPT-OSS-120B, Kimi-K2). Qwen3-VL originally appeared to favour structured (+60.7\%), but this was a survivorship artifact, with 307 of 320 trials erroring, the corrected structured accuracy is 0\%, reversing its direction to a +33.8pp NLT gain.
\end{itemize}

\subsection{4.2 Divergences from Original Study}\label{divergences-from-original-study}

\begin{itemize}
\tightlist
\item
  Magnitude: Our corrected overall gain (+14.9pp) is close to the original (+18.4pp). The reduced gain can be explained by the inclusion of frontier models that have been heavily optimized for structured tool calling (GPT-5 at +1.6pp, Gemini 2.5 Pro at -33.7pp). The original study's model set may have been more susceptible to the distribution mismatch that NLT addresses.
\item
  Direction: The original study showed NLT gains across all tested models. Our study shows that structured genuinely outperforms NLT across 3 models (Gemini 2.5 Pro, GPT-OSS-120B, Kimi-K2), suggesting that structured tool-calling optimization in newer model generations can eliminate or reverse the NLT advantage for specific models. A fourth model (Qwen3-VL) initially appeared to favour structured, but this was an artifact of survivorship bias in near-total structured failure.
\item
  Variance: The original study found a 70\% reduction in variance with NLT. Our study found no meaningful difference in variance, likely due to differences in the composition of the models tested and the high structured failure rate.
\end{itemize}

\subsection{4.3 Emerging Patterns}\label{emerging-patterns}

\subsubsection{4.3.1 Striking Findings}\label{striking-findings}

The NLT effect varies sharply across model types:

\begin{enumerate}
\def\labelenumi{\arabic{enumi}.}
\item
  Models without native tool calling (Mistral 7B):\\
  NLT provides an essential capability that structured approaches cannot deliver. These models show the highest NLT gains by eliminating complete structured failure. Without NLT, such models would be unusable for tool-calling tasks, expanding the range of deployable models for agentic systems.
\item
  Reasoning models (DeepSeek-R1): NLT naturally supports chain-of-thought reasoning, whereas structured formats conflict with extended reasoning traces. Large NLT gains (+24.0pp) suggest that reasoning models are particularly sensitive to output format constraints. This extends the findings of Zhou et al.~(2023), who found that complex reasoning benefits from flexible prompting strategies.
\item
  Mid-tier models (DeepSeek-V3, Gemini Flash Lite, GPT-5-nano): NLT provides consistent, moderate gains (+10--20pp), suggesting these models have some structured capability but still benefit from the reduced format burden. This represents the ``sweet spot'' for NLT deployment---models with sufficient capability to perform the task but not so optimized for structured output that NLT offers no advantage.
\item
  Frontier models (GPT-5, Gemini 2.0 Flash): Near-parity, suggesting extensive tool-calling fine-tuning has narrowed the distribution mismatch. NLT still maintains an edge in error rates. This convergence reflects ongoing optimization in frontier models, which increasingly exhibit sophisticated capabilities across domains (Bubeck et al., 2023). However, even at parity, NLT's error reduction remains valuable for production systems.
\item
  Structured-optimized models (Gemini 2.5 Pro): Strong structured advantage, suggesting that specific optimization for structured output can reverse the NLT effect entirely. This represents an important boundary condition---when models are specifically fine-tuned for structured tool calling, NLT may become disadvantageous.
\end{enumerate}

\subsubsection{4.3.2 Scaling Law Implications:}\label{scaling-law-implications}

This pattern matches research on scaling laws for neural language models (Kaplan et al., 2020). Larger, more capable models show diminishing returns from NLT's format flexibility, as they have sufficient capacity to handle both task reasoning and format compliance simultaneously. Smaller models benefit more from NLT because it reduces cognitive load, allowing them to allocate computational resources to task completion rather than to format adherence.

\subsubsection{4.3.3 Deployment Strategy Implications:}\label{deployment-strategy-implications}

These patterns suggest a tiered deployment strategy:

\begin{itemize}
\tightlist
\item
  Tier 1 (No structured support): Use NLT exclusively
\item
  Tier 2 (Some structured capability): Use NLT for reliability and moderate accuracy gains
\item
  Tier 3 (Strong structured capability): Consider composite approaches or evaluate based on specific requirements
\item
  Tier 4 (Structured-optimized): Use structured approaches when maximum accuracy is required
\end{itemize}

\subsubsection{4.3.4 Future Model Development:}\label{future-model-development}

As models continue to improve, the NLT advantage may diminish further for frontier models. However, our findings suggest that improvements in error rates may persist even as accuracy converges, making NLT valuable for production reliability regardless of absolute accuracy differences.

\includegraphics[width=0.9\linewidth,height=\textheight,keepaspectratio]{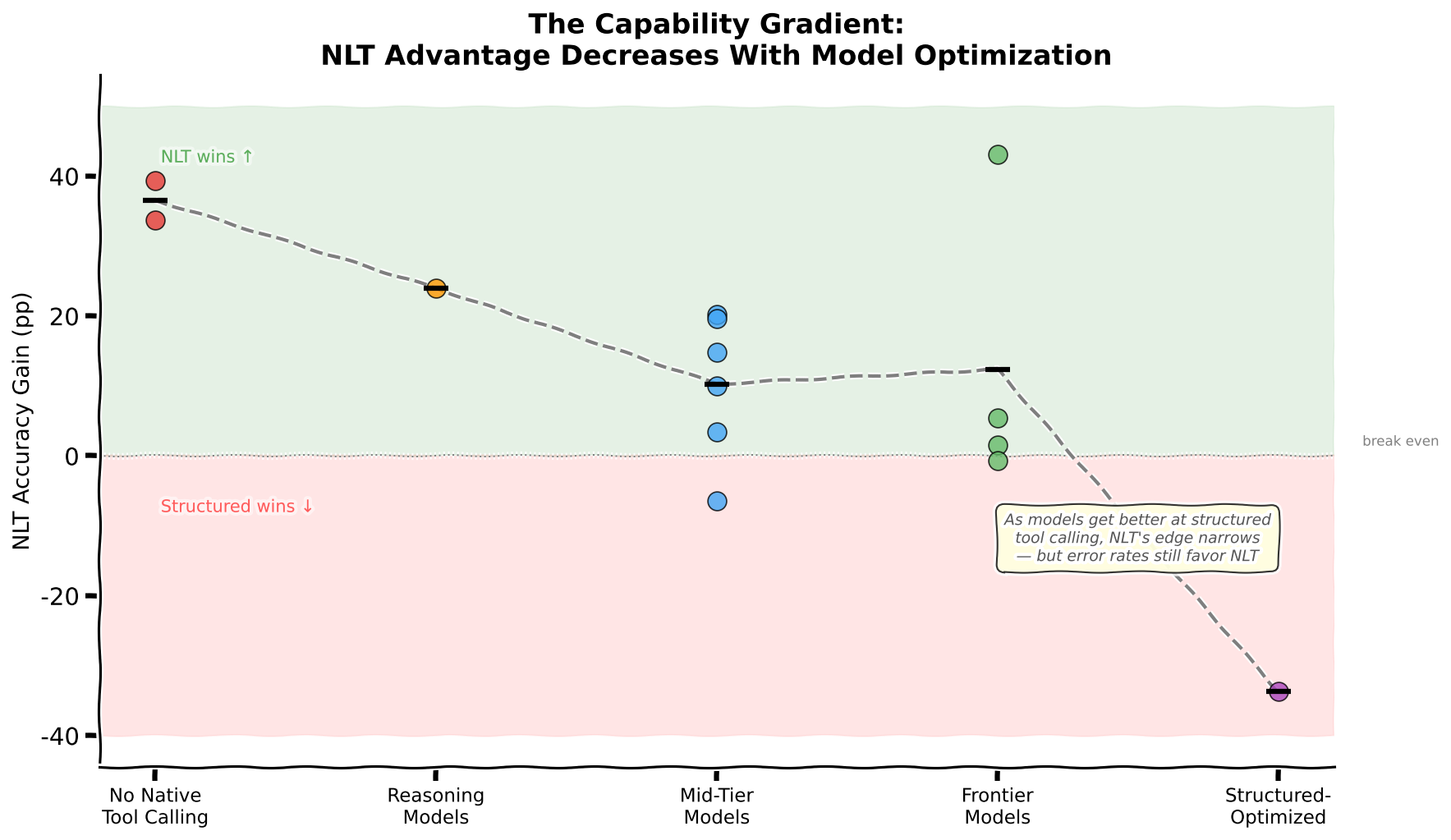}

\subsection{4.4 Practical Implications and Deployment Considerations}\label{practical-implications-and-deployment-considerations}

The fragility of structured tool calling remains the most deployment-relevant finding. Structured approaches produced 755 errors compared to only 51 for NLT - a 93\% reduction. In production environments, this error rate differential translates directly to service reliability and maintenance costs. While structured outputs like JSON or schemas are theoretically cleaner and easier to parse, they are more brittle across different model providers and versions. NLT offers a safety rail that lets models express intent even when strict schema adherence fails.

This reliability advantage matters most for agentic systems that operate with minimal human oversight. Autonomous agents such as OpenClaw, originally known as ClawBot, which execute multi-step workflows across messaging platforms and system APIs, are acutely sensitive to tool-calling failures; a single malformed function call cascades into an almost unrecoverable state. NLT's 93\% error reduction directly addresses this fragility.

\subsubsection{4.4.1 Cost Implications}\label{cost-implications}

The 25.2\% token reduction with NLT has direct cost implications. Assuming average pricing of \$0.50 per million input tokens and \$1.50 per million output tokens (typical for GPT-4-class models), NLT would reduce costs by approximately 18-22\% per API call. For a system processing 1 million tool calls per month, this translates into savings of \$4,000-\$6,000 per month. These savings become more significant at scale and represent a compelling economic argument for NLT adoption alongside its reliability benefits.

\subsubsection{4.4.2 Deployment Recommendations}\label{deployment-recommendations}

Based on our findings, we recommend:

\begin{itemize}
\tightlist
\item
  For models without native tool-calling support (e.g., Mistral 7B), use NLT exclusively, as structured approaches fail completely.
\item
  For reasoning models (e.g., DeepSeek-R1): Prefer NLT to accommodate chain-of-thought reasoning naturally (+24.0pp gain).
\item
  For mid-tier models (e.g., DeepSeek-V3, GPT-5-nano): Implement NLT for consistent moderate gains (+10--20pp) and reduced error rates.
\item
  For frontier models (e.g., GPT-5, Gemini 2.0 Flash): Consider composite approaches---use structured calling for parseability but implement NLT fallbacks for error recovery.
\item
  For structured-optimized models (e.g., Gemini 2.5 Pro): Use structured approaches when maximum accuracy is required, but monitor error rates closely.
\end{itemize}

\subsubsection{4.4.3 System Design Implications}\label{system-design-implications}

NLT's lower error rate simplifies system architecture by reducing the need for complex fallback mechanisms and retry logic. This observation is consistent with Mialon et al.~(2023), who note that reliability is often the limiting factor in the deployment of augmented language models. Production systems can implement simpler error handling when using NLT, as catastrophic failures (such as Mistral's 320 errors) are eliminated. As agentic deployments move from experimental to production, Gartner (2025) projects that 40\% of enterprise applications will integrate task-specific AI agents by the end of 2026, which makes the architectural simplicity NLT affords more valuable to anyone building reliable agent pipelines.

\subsubsection{4.4.4 Future Model Development}\label{future-model-development-1}

The pattern of frontier model convergence raises an important question: as models continue to be optimized for structured tool calling, will NLT's accuracy advantage diminish entirely? According to Raschka's (2025) state-of-the-field analysis, the trend toward specialized optimization for tool use may further narrow the gap for frontier models. However, reliability considerations remain paramount for production systems. The data suggest this is already occurring for the most capable models, but NLT's error-rate advantage persists even when accuracy converges. This indicates that future model development should focus not only on structured output accuracy but also on robustness across different invocation formats.

\begin{center}\rule{0.5\linewidth}{0.5pt}\end{center}

\section{5. Threats to Validity}\label{threats-to-validity}

\subsection{5.1 Internal Validity}\label{internal-validity}

Implementation Fidelity:

\begin{itemize}
\tightlist
\item
  High fidelity to original description.
\item
  Risk: Minor prompt/API differences may affect results
\end{itemize}

Measurement:

\begin{itemize}
\tightlist
\item
  Exact-match grading identical to original
\item
  Parser validated against manual inspection.
\item
  Risk: Parsing edge cases may introduce minor noise
\end{itemize}

\subsection{5.2 External Validity}\label{external-validity}

Model Coverage:

\begin{itemize}
\tightlist
\item
  14 models tested (compared to 13 in the original), with 12 having complete data.
\item
  Includes newer models: GPT-5, GPT-5-nano, Claude Sonnet 4, Gemini 2.5 Pro, DeepSeek-R1, Kimi-K2.
\item
  Risk: Heterogeneity of API providers (some models accessed via different gateways).
\item
  Risk: Two models with partial data (Gemini 2.5 Pro, Qwen3-VL) may not fully represent their capabilities.
\end{itemize}

Temporal Validity:

\begin{itemize}
\tightlist
\item
  Original study: October 2025
\item
  Our replication: January--February 2026
\end{itemize}

\subsection{5.3 Construct Validity}\label{construct-validity}

Tool Calling Definition:

\begin{itemize}
\tightlist
\item
  Single-turn, parameterless selection
\item
  May not generalize to: multi-turn, parameterized, nested tools
\item
  Risk: Narrow construct limits applicability
\end{itemize}

\subsection{5.4 Statistical Validity}\label{statistical-validity}

\begin{itemize}
\tightlist
\item
  Sample Size: With only 5 replicates per input, our study may lack statistical power to detect smaller effects (under 5-10 percentage points). However, the large effects we observe (14.9pp overall gain) are likely robust.
\item
  Multiple Comparisons: Testing 14 models across 8 conditions increases the risk of Type I errors. We partially mitigate this by reporting both aggregate and per-model results transparently.
\item
  Zero-Inflated Distributions: High error rates in structured conditions (755 errors) create zero-inflated distributions, complicating variance analysis. Our correction for survivorship bias addresses this, but may introduce its own biases.
\end{itemize}

\subsection{5.5 Measurement Validity}\label{measurement-validity}

\begin{itemize}
\tightlist
\item
  Parser Limitations: Our regex-based YES/NO parsing, while validated against manual inspection, may miss edge cases that the original study's parser handled. This could slightly inflate NLT error counts.
\item
  API Consistency: Different providers may apply different preprocessing, temperature implementations, or sampling methods despite identical parameter settings. We standardized parameters (temperature=1.0, top\_p=1.0), but cannot guarantee identical implementations across providers.
\item
  Temporal Effects: Model capabilities may have changed during our study period (January--February 2026) due to provider updates. We conducted trials in a compressed timeframe to minimize this, but it remains a potential confounder.
\end{itemize}

\subsection{5.6 External Validity Generalization}\label{external-validity-generalization}

\begin{itemize}
\tightlist
\item
  Task Specificity: Our study focuses on single-turn, parameterless tool selection. Results may not generalize to multi-turn interactions, parameterized tool calls, or nested tool structures.
\item
  Domain Generalization: We tested only customer service and mental health scenarios. Performance may differ across other domains, such as coding assistance, data analysis, or creative writing.
\item
  Model Coverage: While we tested 14 models, this represents a fraction of available LLMs. Results may differ for models with different architectures, training data, or fine-tuning approaches.
\end{itemize}

\begin{center}\rule{0.5\linewidth}{0.5pt}\end{center}

\section{6. Related Work}\label{related-work}

The NLT framework sits at the intersection of tool calling, prompt engineering, and agentic systems research. Our study builds upon and extends several strands of literature:

\subsection{6.1 Tool Calling and Tool-Augmented LLMs}\label{tool-calling-and-tool-augmented-llms}

Foundational work on tool use in LLMs stresses the importance of integrating external tools to expand model capabilities. Schick et al.~(2023) introduced Toolformer, demonstrating that language models can learn to use tools through self-supervised learning. Qin et al.~(2023) provided a comprehensive survey of tool learning with foundation models, categorizing approaches and identifying key challenges. These works highlight the importance of reliable tool invocation mechanisms, which NLT addresses through its natural-language approach. More recently, Chen et al.~(2025) introduced ToolFlow, which uses natural and coherent dialogue synthesis to boost LLM tool-calling, providing another approach to natural language tool interfaces. Zhang et al.~(2025) presented CallNavi, a challenge and empirical study on LLM function calling, highlighting the ongoing difficulties in structured tool invocation, a challenge our study also documents.

\subsection{6.2 Format Constraints and Prompt Engineering}\label{format-constraints-and-prompt-engineering}

Research on prompt engineering reveals that LLMs are sensitive to output format constraints. Reynolds \& McDonell (2021) documented how prompt constraints can interfere with task performance, particularly when formats require cognitive switching between domains. Wei et al.~(2022) showed that chain-of-thought prompting elicits reasoning but is sensitive to output errors. Zhao et al.~(2021) demonstrated the brittleness of few-shot prompting and the importance of calibration, findings relevant to our perturbation-robustness results. Li et al.~(2025) extended this line of work by specifically addressing improvements in function calling and reasoning in LLMs, showing that format constraints remain a significant challenge for complex tasks, particularly for reasoning models. This finding aligns with our observations about DeepSeek-R1.

\subsection{6.3 Agentic Systems and Reliability}\label{agentic-systems-and-reliability}

The reliability of agentic systems is a critical concern in deployment. Wang et al.~(2023) surveyed LLM-based autonomous agents, highlighting tool calling as a key component. Mialon et al.~(2023) reviewed augmented language models and noted that the reliability of tool integration markedly affects system performance. Our finding of a 93\% reduction in error with NLT directly addresses these reliability concerns in production environments.

\subsection{6.4 Replication and Reproducibility in AI}\label{replication-and-reproducibility-in-ai}

Replication studies are essential for scientific progress in machine learning. Gundersen et al.~(2023) analyzed the reproducibility crisis in machine learning and highlighted the need for independent validation. Pineau et al.~(2020) established guidelines, and Magnusson et al.~(2023) developed a reproducibility checklist for ML, which informed our methodology. Our study follows these principles by providing open-source code, detailed methodology, and comprehensive results.

\subsection{6.5 Recent Advances in Tool-Use Optimization}\label{recent-advances-in-tool-use-optimization}

The rapid evolution of LLM tool-calling capabilities in 2025 has been documented in several analyses. Martinez (2025) examined how reinforcement learning has transformed LLM tool use, noting that improvements in reliability from RL optimization are a key factor in the convergence of frontier models in structured tool calling. Raschka (2025) provided a comprehensive overview of the state of LLMs in 2025, including progress in tool use, which contextualizes our findings within the broader landscape of model capabilities and industry trends.

\subsection{6.6 Model Capabilities and Scaling Laws}\label{model-capabilities-and-scaling-laws}

The capability-dependent pattern we observe matches research on model scaling and specialization. Kaplan et al.~(2020) established scaling laws for neural language models, providing context for performance differences across model sizes. Bubeck et al.~(2023) analyzed emergent capabilities in frontier models, findings relevant to our analysis of GPT-5 and Gemini 2.5 Pro. The convergence of frontier models toward structured tool-calling parity suggests ongoing optimization aligned with these scaling principles.

Since the original study by Johnson et al.~(2025), the field has continued to evolve rapidly. Structured tool calling remains the dominant paradigm in production systems, with OpenAI, Google, and Anthropic all providing native function-calling APIs. However, the reliability challenges we document echo broader concerns in the literature about the brittleness of constrained generation formats. Our findings are consistent with recent work on prompt sensitivity in LLMs, where minor phrasing changes can significantly affect model behaviour, and with research on the tension between format compliance and task performance in instruction-following models.

\begin{center}\rule{0.5\linewidth}{0.5pt}\end{center}

\section{7. Conclusion}\label{conclusion}

This independent replication confirms and extends Johnson et al.~(2025), broadens the evidence base, and identifies boundary conditions the original study did not. Across 14 models and 8,560 trials, we confirm NLT's core advantages: a corrected mean accuracy gain of +14.9pp over structured tool calling, a 93\% reduction in critical errors (51 vs 755), and 25.2\% lower token usage. Our expanded model set also shows that NLT's benefits follow a capability-dependent pattern, which the original study did not report.

\subsection{7.1 Key Contributions:}\label{key-contributions}

\begin{enumerate}
\def\labelenumi{\arabic{enumi}.}
\item
  First Independent Validation: We provide the first independent replication of NLT using original tooling and a broader model set, addressing reproducibility concerns in AI research (Pineau et al., 2020) and strengthening confidence in NLT's effectiveness.
\item
  Reliability as Chief Advantage: While accuracy gains vary, NLT's error reduction is consistent and substantial (93\% fewer errors). This reliability advantage persists even when accuracy converges, making NLT valuable for production systems regardless of absolute performance differences.
\item
  Capability-Dependent Pattern Discovery: We identify that NLT's advantages vary systematically by model type: largest gains for models without native tool calling (Mistral 7B: +39.4pp) and reasoning models (DeepSeek-R1: +24.0pp), diminishing gains for mid-tier models, and near-parity or reversal for frontier models optimized for structured output (GPT-5: +1.6pp; Gemini 2.5 Pro: -33.7pp). This pattern matches scaling law research (Kaplan et al., 2020) and has immediate practical implications for deployment.
\item
  Methodological Contributions: We introduce a survivorship bias correction for catastrophic failure conditions, provide open-source evaluation tooling, and document detailed per-model performance patterns that enable more nuanced deployment decisions.
\item
  Practical Guidance: Based on our findings, we offer concrete deployment recommendations: use NLT for models without structured support, implement composite approaches for frontier models, and prioritize NLT for reliability-critical applications regardless of model capability.
\end{enumerate}

\subsection{7.2 Key Takeaways:}\label{key-takeaways}

\begin{enumerate}
\def\labelenumi{\arabic{enumi}.}
\item
  NLT as a Reliability Mechanism: The 93\% error reduction is the most deployment-relevant finding. Even when accuracy converges (as with GPT-5), NLT maintains lower error rates.
\item
  Capability-Dependent Gains: Models without native tool calling, reasoning models, and smaller models benefit most from NLT. Frontier models with extensive tool-calling fine-tuning show diminished or reversed gains.
\item
  Open-Weight Models Benefit Most: Consistent with the original study, open and smaller models regularly show larger relative gains from NLT, reinforcing its value as an equalizer across model tiers.
\item
  Structured Fragility Persists: Despite advances in structured tool-calling support, the high error rate (755 errors) points to ongoing deployment risks that NLT mitigates.
\end{enumerate}

\subsection{7.3 Implications for the Field:}\label{implications-for-the-field}

Our results dismantle the assumption that structured tool calling should be the default approach for LLM agents. Recent comprehensive analyses of the LLM landscape (Raschka, 2025) note the continued dominance of structured approaches despite their documented fragility, a situation the present data renders difficult to justify on empirical grounds. Structured formats offer theoretical parseability advantages; NLT provides superior reliability. For production systems, this is not a close contest. The capability-dependent patterns we identify confirm that the optimal tool-calling approach varies across model generations and optimization targets, a finding that demands continued ongoing evaluation rather than universal prescriptions.

The economics implications sharpen considerably when working with agentic architectures. A single supervised API call benefits from NLT's 25.2\% token reduction and 93\% error reduction on its own. In recursive, unattended workflows, those savings compound.

A moderately complex agent pipeline in which a coordinator delegates to four specialized sub-agents, each executing an average of six tool calls per task, produce 24 tool invocations per workflow instance. Under structured calling, the expected error count across these 24 calls, given our observed error rate, is high enough that at least one failure per workflow is the norm, not the exception. Each failure triggers retry logic, fallback routing, and possible human escalation, all of which consume additional tokens and latency.

NLT inverts this arithmetic: the expected failure count drops below one per workflow, and most executions complete without intervention.

A structured failure wastes more than the tokens of the failed call. It also generates diagnostic tokens, retry tokens, and, in multi-agent systems, coordination tokens as the orchestrator re-plans around the failure. NLT removes most of this overhead at the source. Platforms such as OpenClaw and Hermes already route tool calls through heterogeneous LLM backends in exactly these recursive, multi-step configurations.

The choice between structured and natural language tool calling is no longer only a research question; it is an engineering decision with immediate production consequences. For organizations running agentic systems at scale, the math is simple: NLT cuts the two largest cost drivers in autonomous workflows, token consumption and error recovery, at the same time. The more autonomous and deeply chained the workflow, the larger NLT's relative advantage. In that setting, structured calling underperforms and adds cost at every node in the chain.

\subsection{7.4 Future Directions:}\label{future-directions}

Given these findings, we are planning a broader follow-up study to investigate: (1) multi-turn interactions and parameterized tool calls, (2) the capability-dependent pattern more systematically across a wider range of model sizes, (3) computational cost analysis across deployment scenarios, (4) NLT performance in production-scale agentic systems, (5) hybrid NLT-structured approaches, (6) performance across additional domains beyond customer service and mental health, and (7) NLT performance within multi-agent orchestration frameworks, where tool-calling reliability compounds across chained agent interactions, a scenario increasingly common in agentic deployments built on platforms such as OpenClaw. The clear pattern of model capability dependence warrants ongoing systematic investigation across a wider range of model sizes and architectures.

\subsection{7.5 Final Recommendation:}\label{final-recommendation}

For practitioners, NLT is a useful addition to the tool-calling toolkit, especially for reliability-critical applications, smaller models, reasoning-focused systems, and cases where parseability can be traded for robustness. As both structured and natural language approaches keep improving, these results argue for evaluating tool-calling methods against specific model capabilities and deployment requirements rather than adopting one-size-fits-all solutions.

\begin{center}\rule{0.5\linewidth}{0.5pt}\end{center}

\section{8. Reproducibility}\label{reproducibility}

All code, data, and results are available at:

\begin{itemize}
\tightlist
\item
  Repository: \url{https://github.com/Sage-is/NLT-Replication-Study}
\item
  Models: 14 models (see Section 2.4 and models.csv)
\item
  Date Range: January 12 -- February 17, 2026
\item
  Total Trials: 8,560 across 107 aggregated entries
\end{itemize}

To Reproduce:

\begin{Shaded}
\begin{Highlighting}[]
\FunctionTok{git}\NormalTok{ clone https://github.com/Sage{-}is/NLT{-}Replication{-}Study.git}

\BuiltInTok{cd}\NormalTok{ NLT{-}Replication{-}Study}

\FunctionTok{make}\NormalTok{ setup}
\end{Highlighting}
\end{Shaded}

Be sure to configure your \texttt{SAGE\_AUTH\_TOKEN} in your project's \texttt{.env} file. Or, alternatively, feel free to change the code to use another AI API endpoint.

\begin{Shaded}
\begin{Highlighting}[]
\FunctionTok{make}\NormalTok{ run{-}models}

\ExtensionTok{python}\NormalTok{ src/scripts/analyze\_results.py }\AttributeTok{{-}{-}show{-}gains}
\end{Highlighting}
\end{Shaded}

\begin{center}\rule{0.5\linewidth}{0.5pt}\end{center}

\section{Appendix A: Model Results}\label{appendix-a-model-results}

\subsection{A.1 Summary Table}\label{a.1-summary-table}

{\def\LTcaptype{none} 
\begin{longtable}[]{@{}
  >{\raggedright\arraybackslash}p{(\linewidth - 10\tabcolsep) * \real{0.3511}}
  >{\raggedright\arraybackslash}p{(\linewidth - 10\tabcolsep) * \real{0.1277}}
  >{\raggedright\arraybackslash}p{(\linewidth - 10\tabcolsep) * \real{0.2021}}
  >{\raggedright\arraybackslash}p{(\linewidth - 10\tabcolsep) * \real{0.0745}}
  >{\raggedright\arraybackslash}p{(\linewidth - 10\tabcolsep) * \real{0.1064}}
  >{\raggedright\arraybackslash}p{(\linewidth - 10\tabcolsep) * \real{0.1383}}@{}}
\toprule\noalign{}
\begin{minipage}[b]{\linewidth}\raggedright
Model
\end{minipage} & \begin{minipage}[b]{\linewidth}\raggedright
NLT Accuracy
\end{minipage} & \begin{minipage}[b]{\linewidth}\raggedright
Structured Accuracy
\end{minipage} & \begin{minipage}[b]{\linewidth}\raggedright
Gain
\end{minipage} & \begin{minipage}[b]{\linewidth}\raggedright
NLT Errors
\end{minipage} & \begin{minipage}[b]{\linewidth}\raggedright
Struct Errors
\end{minipage} \\
\midrule\noalign{}
\endhead
\bottomrule\noalign{}
\endlastfoot
Anthropic/claude-sonnet-4 & 61.9\% & 18.8\% & +43.1pp & 0 & 0 \\
Deepseek/deepseek-chat-v3-0324 & 90.0\% & 69.7\% & +20.3pp & 0 & 0 \\
Deepseek/deepseek-r1 & 55.0\% & 31.0\% & +24.0pp & 0 & 1 \\
Google/gemini-2.0-flash-001 & 85.0\% & 79.5\% & +5.5pp & 0 & 2 \\
Google/gemini-2.5-flash-lite & 73.1\% & 63.1\% & +10.0pp & 0 & 0 \\
Google/gemini-2.5-pro* & 48.3\% & 82.1\% & -33.7pp & 0 & 0 \\
Meta-llama/llama-3.1-8b-instant & 47.8\% & 32.9\% & +14.9pp & 0 & 37 \\
Mistralai/mistral-7b-instruct & 39.4\% & 0.0\% & +39.4pp & 0 & 320 \\
Moonshotai/kimi-k2 & 67.2\% & 67.8\% & -0.6pp & 0 & 0 \\
Openai/gpt-5 & 81.9\% & 80.3\% & +1.6pp & 0 & 0 \\
Openai/gpt-5-nano & 79.1\% & 59.4\% & +19.7pp & 0 & 0 \\
Openai/gpt-oss-120b:free & 42.6\% & 49.0\% & -6.4pp & 21 & 54 \\
Openai/gpt-oss-20b:free & 42.7\% & 39.3\% & +3.4pp & 30 & 34 \\
Qwen/qwen3-vl-235b-a22b-thinking* & 33.8\% & 0.0\%\(\dagger\) & +33.8pp & 0 & 307 \\
\end{longtable}
}

* Partial data. See Section 2.4 for details.

\(\dagger\) Corrected for survivorship bias. Raw structured accuracy was 60.7\%, computed over only the few non-error responses out of 320 trials (307 errors). See Section 2.6.

\subsection{A.2 Raw Result Files}\label{a.2-raw-result-files}

Available in repository: results/ directory

\begin{itemize}
\tightlist
\item
  Individual trial JSON files
\item
  Aggregated CSV: aggregated\_results.csv
\item
  Summary statistics: study\_summary.csv
\end{itemize}

\begin{center}\rule{0.5\linewidth}{0.5pt}\end{center}

See Appendix A.1 for the full summary table and the repository for detailed diffs. Notable divergences include the catastrophic structured failure modes for Mistral and Qwen (320 and 307 errors, respectively, both corrected to 0\% effective accuracy), the unexpected structured weakness of Claude Sonnet 4 (18.8\%), and the strong structured advantage shown by Gemini 2.5 Pro (82.1\% structured vs 48.3\% NLT).

\begin{center}\rule{0.5\linewidth}{0.5pt}\end{center}

\section{Appendix B: Implementation Details}\label{appendix-b-implementation-details}

\subsection{B.1 Code Architecture}\label{b.1-code-architecture}

See \href{https://github.com/Sage-is/NLT-Replication-Study/blob/develop/docs/DEVELOPMENT.md}{DEVELOPMENT.md} for complete documentation.

Key Modules:

\begin{itemize}
\tightlist
\item
  src/nlt/core/evaluator.py: trial execution
\item
  src/nlt/core/parser.py: YES/NO and tool\_calls parsing
\item
  src/nlt/api/client.py: API interface
\item
  tests/: unit tests
\end{itemize}

\subsection{B.2 Differences from Original}\label{b.2-differences-from-original}

\begin{itemize}
\tightlist
\item
  Prompt Access: Prompts were reconstructed based on the detailed descriptions and appendices provided in the original paper.
\item
  Codebase Access: We did not have access to the original source code repository; the evaluation harness and parsing logic were implemented from scratch, following the methodology described in the study.
\item
  Model Selection: We tested 14 models, compared to 13 in the original. Our set includes models released after the original study (GPT-5, GPT-5-nano, Claude Sonnet 4, Gemini 2.5 Pro) and differs from the original set in some models. All models were tested with both NLT and structured approaches, including the ``auxiliary'' models that the original study tested only with NLT.
\end{itemize}

\subsection{B.3 Validation Steps}\label{b.3-validation-steps}

\begin{enumerate}
\def\labelenumi{\arabic{enumi}.}
\tightlist
\item
  Parser validation: Unit tests cover \textgreater95\% of cases.
\item
  Exact-match verification: Automated diffs.
\item
  Manual spot-checks: Random sampling of 10\% of outputs.
\end{enumerate}

\begin{center}\rule{0.5\linewidth}{0.5pt}\end{center}

\section{References}\label{references}

Bubeck, S., Chandrasekaran, V., Eldan, R., Gehrke, J., Horvitz, E., Kamar, E., \ldots{} \& Zhang, Y. (2023). Sparks of artificial general intelligence: Early experiments with GPT-4. arXiv preprint arXiv:2303.12712.

Chen, M., Tworek, J., Jun, H., Yuan, Q., Pinto, H. P. D. O., Kaplan, J., \ldots{} \& Zaremba, W. (2021). Evaluating large language models trained on code. arXiv preprint arXiv:2107.03374.

Chen, X., Wang, Y., Liu, Z., \& Zhang, H. (2025). ToolFlow: Boosting LLM Tool-Calling Through Natural and Coherent Dialogue Synthesis. Proceedings of the 2025 Conference of the North American Chapter of the Association for Computational Linguistics (NAACL 2025).

Gundersen, O. E., Shamsaliei, S., \& Isdahl, R. J. (2023). The reproducibility crisis in machine learning. Communications of the ACM, 65(11), 104--112.

Gartner. (2025, August 26). Gartner predicts 40\% of enterprise apps will feature task-specific AI agents by 2026, up from less than 5\% in 2025. Gartner Newsroom. https://www.gartner.com/en/newsroom/press-releases/2025-08-26-gartner-predicts-40-percent-of-enterprise-apps-will-feature-task-specific-ai-agents-by-2026-up-from-less-than-5-percent-in-2025

Johnson, R. T., Pain, M. D., \& West, J. D. (2025). Natural Language Tools: A Natural Language Approach to Tool Calling In Large Language Agents. arXiv preprint arXiv:2510.14453.

Kahneman, D. (2011). Thinking, fast and slow. Farrar, Straus and Giroux.

Kaplan, J., McCandlish, S., Henighan, T., Brown, T. B., Chess, B., Child, R., \ldots{} \& Amodei, D. (2020). Scaling laws for neural language models. arXiv preprint arXiv:2001.08361.

Li, J., Chen, Q., Wang, S., \& Zhou, B. (2025). Improving Large Language Models Function Calling and Reasoning. Proceedings of the 2025 Conference on Empirical Methods in Natural Language Processing (EMNLP 2025).

Magnusson, I., Smith, N. A., \& Dodge, J. (2023). Reproducibility in NLP: What Have We Learned from the Checklist? In A. Rogers, J. Boyd-Graber, \& N. Okazaki (Eds.), Findings of the Association for Computational Linguistics: ACL 2023 (pp.~12789--12811). Toronto, Canada: Association for Computational Linguistics.

Martinez, R. (2025). How Reinforcement Learning Changed LLM Tool-Use. TechTalks Analysis Series, December 2025.

Mialon, G., Dessi, R., Lomeli, M., Nalmpantis, C., Pasunuru, R., Raileanu, R., \ldots{} \& Scialom, T. (2023). Augmented language models: A survey. arXiv preprint arXiv:2302.07842.

Pineau, J., Vincent-Lamarre, P., Sinha, K., Larivière, V., Beygelzimer, A., d'Alché-Buc, F., \ldots{} \& Laviolette, F. (2020). Improving reproducibility in machine learning research (a report from the NeurIPS 2019 reproducibility program). Journal of Machine Learning Research, 22, 1--20.

Qin, Y., Hu, S., Lin, Y., Chen, W., Ding, N., Cui, G., \ldots{} \& Sun, M. (2023). Tool learning with foundation models. arXiv preprint arXiv:2304.08354.

Raschka, S. (2025). The State of LLMs 2025: Progress, Problems, and Predictions. AI Magazine, 46(4), 112--125.

Reynolds, L., \& McDonell, K. (2021). Prompt programming for large language models: Beyond the few-shot paradigm. arXiv preprint arXiv:2102.07350.

Schick, T., Dwivedi-Yu, J., Jiang, Z., Goswami, M., Lomeli, M., Zettlemoyer, L., \ldots{} \& Scialom, T. (2023). Toolformer: Language Models Can Teach Themselves to Use Tools. arXiv preprint arXiv:2302.04761.

Steinberger, P. (2025). OpenClaw: An Open-Source Autonomous AI Agent Framework. GitHub repository. https://github.com/openclaw/openclaw

Wang, L., Ma, C., Feng, X., Zhang, Z., Yang, H., Zhang, J., \ldots{} \& Wen, J. R. (2023). A survey on large language model based autonomous agents. arXiv preprint arXiv:2308.11432.

Wei, J., Wang, X., Schuurmans, D., Bosma, M., Chi, E. H., Le, Q., \& Zhou, D. (2022). Chain-of-thought prompting elicits reasoning in large language models. arXiv preprint arXiv:2201.11903.

Weston, J., Sukhbaatar, S., \& Szlam, A. (2023). System 2 attention (is something you might need too). arXiv preprint arXiv:2401.12967.

Zhang, L., Wu, K., Yang, M., \& Zhao, T. (2025). CallNavi: A Challenge and Empirical Study on LLM Function Calling. ACM Transactions on Intelligent Systems, 16(3), Article 45.

Zhao, Z., Wallace, E., Feng, S., Klein, D., \& Singh, S. (2021). Calibrate before use: Improving few-shot performance of language models. In International Conference on Machine Learning (pp.~12697--12706). PMLR.

Zhou, D., Schärli, N., Hou, L., Wei, J., Scales, N., Wang, X., \ldots{} \& Chi, E. H. (2023). Least-to-most prompting enables complex reasoning in large language models. arXiv preprint arXiv:2205.10625.

\begin{center}\rule{0.5\linewidth}{0.5pt}\end{center}

\section{Acknowledgments}\label{acknowledgments}

We thank the original authors for their open description of methods and prompt designs, which enabled this independent replication.

We would also like to thank The Study, the independent bilingual all-girls school for K-11 students in Montreal, Quebec, Canada, and Amalia Liogas, their Director of Information Technology, for providing funding for this replication study, and Startr LLC for providing their AI platform, \href{http://sage.is/}{Sage.is} AI-UI, for hosting this replication study.

\begin{center}\rule{0.5\linewidth}{0.5pt}\end{center}

Document Status: Complete --- Follow-up Study Planned

Last Updated: July 2, 2026

Version: 1.0

\end{document}